\documentclass[11pt,a4paper]{article}
\usepackage[hyperref]{naaclhlt2018}
\usepackage{times}
\usepackage{latexsym}
\usepackage{subfigure}
\usepackage{graphicx}
\usepackage{multirow}
\usepackage{url}
\usepackage{amsmath}
\usepackage{color}
\usepackage{float}
\usepackage{booktabs}
\usepackage{amssymb}
\usepackage{capt-of}
\usepackage{cleveref}
\crefname{section}{§}{§§}
\Crefname{section}{§}{§§}

\definecolor{red}{rgb}{0.8, 0.0, 0.0}
\definecolor{green}{rgb}{0.0, 0.5, 0.0}

\usepackage[english]{babel}
\usepackage[utf8]{inputenc}
\usepackage{algorithm}
\usepackage[noend]{algpseudocode}

\aclfinalcopy 

\title{Towards a Continuous Knowledge Learning Engine for Chatbots}

\author{Sahisnu Mazumder,~~ Nianzu Ma,~~ Bing Liu\\ 
	Department of Computer Science, 
	University of Illinois at Chicago, IL, USA\\
	sahisnumazumder@gmail.com,~nma4@uic.edu,~liub@cs.uic.edu}

\date{}

\begin{document}
\maketitle

\begin{abstract}
Although chatbots have been very popular in recent years, they still have some serious weaknesses which limit the scope of their applications. One major weakness is that they cannot learn new knowledge during the conversation process, i.e., their knowledge is fixed beforehand and cannot be expanded or updated during conversation. In this paper, we propose to build a general knowledge learning engine for chatbots to enable them to continuously and interactively learn new knowledge during conversations. As time goes by, they become more and more knowledgeable and better and better at learning and conversation. We model the task as an \textit{open-world knowledge base completion} problem and propose a novel technique called \textit{lifelong interactive learning and inference} (LiLi) to solve it. LiLi works by imitating how humans acquire knowledge and perform inference during an interactive conversation. Our experimental results show LiLi is highly promising. 

\end{abstract}

\section{Introduction}

Chatbots such as dialog and question-answering systems have a long history in AI and natural language processing. Early such systems were mostly built 
using markup languages such as AIML\footnote{http://www.alicebot.org/}, handcrafted conversation
generation rules, and/or information retrieval techniques \cite{banchs2012iris,ameixa2014luke,lowe2015ubuntu,serban2015survey}. 
Recent neural conversation models \cite{vinyals2015neural,xing2017topic,li2017adversarial} are even able to perform open-ended conversations. However, since they do not use explicit knowledge bases and do not perform inference, they often suffer from generic and dull responses \cite{xing2017topic,li2017data}. More recently,  \citet{ghazvininejad2017knowledge} and \citet{le2016lstm} proposed to use knowledge bases (KBs) to help generate responses for knowledge-grounded conversation.  However, one major weakness of all existing chat systems is that they do not explicitly or implicitly learn new knowledge in the conversation process. This seriously limits the scope of their applications. In contrast, we humans constantly learn new knowledge in our conversations. Even if some existing systems can use very large knowledge bases either harvested from a large data source such as the Web or built manually, these KBs still miss a large number of facts (knowledge) \cite{west2014knowledge}. It is thus important for a chatbot to continuously learn new knowledge in the conversation process to expand its KB and to improve its conversation ability. 

In recent years, researchers have studied the problem of \textit{KB completion}, i.e., inferring new facts (knowledge) automatically from existing facts in a KB. KB completion (KBC) is defined as a binary classification problem: Given a query triple, ($s$, $r$, $t$), we want to predict whether the source entity $s$ and target entity $t$ can be linked by the relation $r$. However, existing approaches \cite{lao2011random,lao2015learning,bordes2011learning,bordes2013translating,nickel2015review,mazumder2017context} solve this problem under the \textit{closed-world} assumption, i.e., $s$, $r$ and $t$ are all \textit{known} to exist in the KB. This is a major weakness because it means that no new knowledge or facts may contain unknown entities or relations. Due to this limitation, KBC is clearly not sufficient for knowledge learning in conversations because in a conversation, the user can say anything, which may contain entities and relations that are not already in the KB. 

In this paper, we remove this assumption of KBC, and allow all $s$, $r$ and $t$ to be \textit{unknown}.  We call the new problem  \textit{open-world knowledge base completion} (OKBC). OKBC generalizes KBC. Below, we show that 
solving OKBC \textit{naturally} provides the ground for knowledge learning and inference in conversations. In essence, we formulate an \textit{abstract} problem of knowledge learning and inference in conversations as a \textit{well-defined} OKBC problem in the interactive setting. 

From the perspective of knowledge learning in conversations, essentially we can extract two key types of information, true facts and queries, from the user utterances. Queries are facts whose truth values need to be determined\footnote{In this work we do not deal with subjective information such as beliefs and opinions, which we leave it to future work.}. Note that we do not study fact or relation extraction in this paper as there is an extensive work on the topic.  (1) For a true fact, we will incorporate it into the KB. Here we need to make sure that it is not already in the KB, which involves relation resolution and entity linking.  After a fact is added to the KB, we may predict that some related facts involving some existing relations in the KB may also be true (not logical implications as they can be automatically inferred). For  example, if the user says ``Obama was born in USA,'' the system may guess that (\textit{Obama}, \textit{CitizenOf}, \textit{USA}) (meaning that Obama is a citizen of USA) could also be true based on the current KB. To verify this fact, it needs to solve a KBC problem by treating (\textit{Obama}, \textit{CitizenOf}, \textit{USA}) as a query. This is a KBC problem because the fact (\textit{Obama}, \textit{BornIn}, \textit{USA}) extracted from the original sentence has been added to the KB. Then Obama and USA are in the KB. If the KBC problem is solved, it learns a new fact (Obama, CitizenOf, USA) in addition to the extracted fact (Obama, BornIn, USA). 
(2) For a query fact, e.g., (Obama, BornIn, USA) extracted from the user question ``Was Obama born in USA?'' we need to solve the OKBC problem if any of ``\textit{Obama}, ``\textit{BornIn}'', or ``\textit{USA}" is not already in the KB. 

We can see that OKBC is the core of a knowledge learning engine for conversation. Thus, in this paper, we focus on solving it. We assume that other tasks such as fact/relation extraction and resolution and guessing of related facts of an extracted fact are solved by other sub-systems.

We solve the OKBC problem by mimicking how humans acquire knowledge and perform reasoning in an interactive conversation. Whenever we encounter an unknown concept or relation while answering a query, we perform inference using our existing knowledge. If our knowledge does not allow us to draw a conclusion, we typically ask questions to others to acquire related knowledge and use it in inference. The process typically involves an \textit{inference strategy} (a sequence of actions), which interleaves a sequence of \textit{processing} and \textit{interactive} actions. A processing action can be the selection of related facts, deriving inference chain, etc., that advances the inference process. An interactive action can be deciding what to ask, formulating a suitable question, etc., that enable us to interact. The process helps grow the knowledge over time and the gained knowledge enables us to communicate better in the future. We call this \textbf{lifelong interactive learning and inference} (LiLi). 
Lifelong learning is reflected by the facts that the newly acquired facts are retained in the KB and used in inference for future queries, and that the accumulated knowledge in addition to the updated KB including past inference performances are leveraged to guide future interaction and learning. LiLi should have the following capabilities: 

\begin{enumerate}
	\vspace{-0.1cm}
	\item to \textbf{\textit{formulate an inference strategy}} for a given query that embeds \textit{processing} and \textit{interactive} actions.
	\vspace{-0.1cm}
	\item to \textbf{\textit{learn interaction behaviors}} (deciding what to ask and when to ask the user). 
	\vspace{-0.1cm}
	\item to \textbf{\textit{leverage the acquired knowledge}} in the current and future inference process.
	\vspace{-0.1cm}
	\item to \textbf{\textit{perform 1, 2 and 3 in a lifelong manner}} for continuous knowledge learning.
	\vspace{-0.1cm}
\end{enumerate}

This setting is ideal for many NLP applications like dialog and question-answering systems that naturally provide the scope for human interaction and demand real-time inference.

LiLi starts with the closed-world KBC approach \textit{path-ranking} (PR) 
\cite{lao2011random,gardner2015efficient}\footnote{We choose path-ranking (PR) due to its high interpretability and better performance than latent features \cite{wang2016knowledge,toutanova2015observed}} 
and extends KBC in a major way to open-world knowledge base completion (OKBC).  
For a relation $r$, PR works by enumerating paths (except single-link path $r$) between entity-pairs linked by $r$ in the KB and use them as features to train a binary classifier to predict whether a query $(s, r, t)$ should be in the KB. Here, a path between two entities is a sequence of relations linking them. In our work, we adopt the latest PR method, C-PR \cite{mazumder2017context} and extend it to make it work in the open-world setting. C-PR enumerates paths by performing bidirectional random walks over the KB graph while leveraging the context of the source-target entity-pair. We also adopt and extend the compositional vector space model \cite{neelakantan2015compositional,das2016chains} with continual learning capability for prediction. 

\begin{table}[t]
	\tiny
	\centering
	\label{my-label}
	\small
	\begin{tabular}{lp{6cm}}
		\textbf{\small User:} &  (Obama, CitizenOf, USA)? ``Is Obama a citizen of USA?'' \hfill[\textcolor{red}{\textbf{Query}}] \\
		\textbf{\small LiLi:} & I do not know what ``\textbf{CitizenOf}'' mean? Can you provide me an example? \hfill[\textit{Ask for Clue}] \\
		\textbf{\small User:} & (David Cameron, CitizenOf, UK). ``David Cameron is a citizen of UK.''
		\hfill[\textcolor{green}{\textbf{SF1}}]\\
		\textbf{\small LiLi:} & Got it. Can you tell me how ``\textbf{Obama}'' and ``\textbf{Honolulu}'' are related? \hfill[\textit{CLQ}] \\
		\textbf{\small User:} & (Obama, BornIn, Honolulu). ``Obama was born in Honolulu.''\hfill[\textcolor{green}{\textbf{SF2}}]\\
		\textbf{\small LiLi:}  & Got it. Can you tell also me how ``\textbf{Honolulu}'' and ``\textbf{Hawaii}'' are related? \hfill[\textit{MLQ}] \\
		\textbf{\small User:} & (Honolulu, CapitalOfState, Hawaii). ``Honolulu is the state capital of Hawaii.'' \hfill[\textcolor{green}{\textbf{SF3}}]\\
		\textbf{\small LiLi:} & (Obama, CitizenOf, USA) is correct. \hfill[\textcolor{blue}{\textbf{Answer}}]\\
	\end{tabular}
	\captionof{figure}{\small An example of interactive inference and learning. Note that LiLi only works with triples. Each triple above is assumed to be extracted from the sentence after it. }
	\normalsize
	\vspace{-0.3cm}
\end{table}

Given an OKBC query ($s$, $r$, $t$) (e.g., ``(\textit{Obama}, \textit{CitizenOf}, \textit{USA}), which means whether Obama a citizen of USA), 
LiLi interacts with the user (\textit{if needed}) by \textit{dynamically} formulating questions (see the interaction example in Figure 1\footnote{Note that the user query and responses are in triples as we are not building a conversation system but a knowledge acquisition system. Also, the query may be from a user or a system (e.g., a question-answer system, a conversation system that has extracted a candidate fact and wants to verify it and add it to the KB. This paper will not study the case that the query fact is already in the KB, which is easy to verify. Also note that, as our work focuses on knowledge learning and inference, \textit{rather than conversation modeling}, we simply use template-based question generation to model LiLi's interaction with the user.}, which will be further explained in \S3) and leverages the interactively acquired knowledge (supporting facts (SFs) in the figure) for continued inference.
To do so, LiLi formulates a query-specific inference strategy and executes it. We design LiLi in a Reinforcement Learning (RL) setting that performs sub-tasks like formulating and executing strategy, training a prediction model for inference, and knowledge retention for future use. To the best of our knowledge, our work is the first to address the OKBC problem and to propose an interactive learning mechanism to solve it in a continuous or lifelong manner. We empirically verify the effectiveness of LiLi on two \textit{standard} real-world KBs: \textit{Freebase} and \textit{WordNet}. Experimental results show that LiLi is highly effective in terms of its predictive performance and strategy formulation ability.

\section{Related Work} 
\label{sec:RW}
To the best of our knowledge, we are not aware of any knowledge learning system that can learn new knowledge in the conversation process. This section thus discusses other related work. 

Among existing KB completion approaches, \cite{neelakantan2015compositional} extended the vector space model for zero-shot KB inference. However, the model cannot handle unknown entities and can only work on fixed set of unknown relations with known embeddings. Recently, \cite{shi2017open} proposed a method using external text corpus to perform inference on unknown entities. However, the method cannot handle unknown relations. Thus, these methods are not suitable for our open-world setting. None of the existing KB inference methods perform interactive knowledge learning like LiLi. 

NELL \cite{mitchell2015never} continuously updates its KB using facts extracted from the Web. Our task is very different as we do not do Web fact extraction (which is also useful). We focus on user interactions in this paper. 

Our work is related to interactive language learning (ILL) \cite{wang2016learning,wang2017naturalizing}, but these are not about KB completion. The work in \cite{li2016learning} allows a learner to ask questions in dialogue. However, this work used RL to learn about \textit{whether to ask the user or not}. The ``\textit{what to ask} aspect" was manually designed by modeling synthetic tasks. LiLi formulates query-specific inference strategies which embed interaction behaviors. Also, no existing dialogue systems \cite{vinyals2015neural,li2016dialogue,bordes2016learning,weston2016dialog,zhang2017listen} employ lifelong learning to train prediction models by using information/knowledge retained in the past. 

Our work is related to general lifelong learning in~\cite{ChenAndLiubook2016,Ruvolo2013ELLA,Chen2014ICML,Chen2015ACL,BouAmmar2015CrossDomainLRL,Lei2017lifelong}. However, they learn only one type of tasks, e.g., supervised, topic modeling or reinforcement learning (RL) tasks. None of them is suitable for our setting, which involves interleaving of RL, supervised and interactive learning. More details about lifelong learning can be found in the book~\cite{ChenAndLiubook2016}.

\section{Interactive Knowledge Learning (LiLi)}
\label{sec:LiLi}
We design LiLi as a combination of two interconnected models: (1) a RL model that learns to formulate a query-specific inference strategy for performing the OKBC task, and (2) a lifelong prediction model to predict whether a triple should be in the KB, which is invoked by an action while executing the inference strategy and is learned for each relation as in C-PR. The framework improves its performance over time through user interaction and knowledge retention. Compared to the existing KB inference methods, LiLi overcomes the following three challenges for OKBC:

1. \textbf{Mapping open-world to close-world}. Being a closed-world method,  C-PR cannot extract path features and learn a prediction model when any of $s$, $r$ or $t$ is unknown. LiLi solves this problem through \textit{interactive} knowledge acquisition. If $r$ is \textit{unknown}, LiLi asks the user to provide a clue (an example of $r$). And if $s$ or $t$ is \textit{unknown}, LiLi asks the user to provide a link (relation) to connect the unknown entity with an existing entity (automatically selected) in the KB. We refer to such a query as a \textit{connecting link query} (CLQ). The acquired knowledge reduces OKBC to KBC and makes the inference task feasible. 

2. \textbf{Spareseness of KB.} A main issue of all PR methods like C-PR is the connectivity of the KB graph. If there is no path connecting $s$ and $t$ in the graph, path enumeration of C-PR gets stuck and inference becomes infeasible. In such cases, LiLi uses a template relation (``@-?-@") as the \textit{missing link} marker to connect entity-pairs and continues feature extraction. A path containing ``@-?-@" is called an \textit{incomplete path}. Thus, the extracted feature set contains both complete (no missing link) and incomplete paths. Next, LiLi selects an incomplete path from the feature set and asks the user to provide a link for path completion. We refer to such a query as \textit{missing link query} (MLQ).

3. \textbf{Limitation in user knowledge.} If the user is unable to respond to MLQs or CLQs, LiLi uses a \textit{guessing mechanism} (discussed later) to fill the gap. This enables LiLi to continue its inference even if the user cannot answer a system question. 

\subsection{Components of LiLi}
As lifelong learning needs to retain knowledge learned from past tasks and use it to help future learning \cite{ChenAndLiubook2016}, LiLi uses a \textbf{Knowledge Store (KS)} for knowledge retention. KS has four components: \textbf{(i) Knowledge Graph} ($G$): $G$ (the KB) is initialized with base KB triples (see \S4) and gets updated over time with the acquired knowledge.
\textbf{(ii) Relation-Entity Matrix} ($\mathcal{M}$): $\mathcal{M}$ is a sparse matrix, with rows as relations and columns as entity-pairs and is used by the prediction model. Given a triple ($s$, $r$, $t$) $\in G$, we set $\mathcal{M}$[$r$, ($s$, $t$)] = 1 indicating $r$ occurs for pair ($s$, $t$). \textbf{(iii) Task Experience Store} ($\mathcal{T}$): $\mathcal{T}$ stores the predictive performance of LiLi on past learned tasks in terms of \textit{Matthews correlation coefficient} (MCC)\footnote{ en.wikipedia.org/wiki/Matthews\_correlation\_coefficient} that measures the quality of binary classification. So, for two tasks $r$ and $r'$ (each relation is a task), if $\mathcal{T}$[$r$] $>$ $\mathcal{T}$[$r'$] [where $\mathcal{T}$[$r$]=MCC($r$)], we say C-PR has learned $r$ well compared to $r'$. \textbf{(iv) Incomplete Feature DB} ($\Pi_{DB}$): $\Pi_{DB}$ stores the frequency of an incomplete path $\pi$ in the form of a tuple ($r$, $\pi$, $e_{ij}^\pi$) and is used in formulating MLQs. $\Pi_{DB}$[($r$, $\pi$, $e_{ij}^\pi$)] = $N$ implies LiLi has extracted incomplete path $\pi$ $N$ times involving entity-pair $e_{ij}^\pi$ [($e_i$, $e_j$)] for query relation $r$.

\begin{table*}[t]
	\begin{minipage}{.39\textwidth}
		\tiny
		\centering
		\caption{\small Parameters of LiLi.}
		\vspace{0.1cm}
		\label{my-label}
		\begin{tabular}{|c|p{5.25cm}|}
			\hline
			\multicolumn{1}{|c|}{\textbf{Para.}} & \multicolumn{1}{c|}{\textbf{Description}}                                           \\ \hline
			$\alpha$ 	& learning rate of Q-learning agent        \\ \hline
			$\gamma$ 	& discount factor of Q-learning agent     \\ \hline
			$\delta_{\pi}$    & If $S_t$[ILO]=0 and feature set contains  $\geq \delta_{\pi}$ \# complete features, we consider feature set as \textit{complete} and set $S_t$[CPF]=1.    \\  \hline
			$\delta_{IL}$	 & max \# times LiLi is allowed to ask user per query (we refer $\delta_{IL}$ as the \textit{interaction limit} of LiLi per query). \\ \hline
			$\eta_{\pi}$    & maximum path length for C-PR    \\  \hline
			$\eta_{w}$ & \begin{tabular}[c]{@{}l@{}}number of random walks per query for C-PR \end{tabular}    \\ \hline
			$l,~h$  & low and high contextual similarity threshold   \\ \hline
			$k$   & rank of trancated SVD     \\ \hline
			$\beta$    & clue acquisition rate                                                                          \\ \hline
			$\rho$   & \begin{tabular}[c]{@{}l@{}}past task selection rate 
			\end{tabular} \\ \hline
		\end{tabular}
		\normalsize
		\vspace{-0.1cm}
	\end{minipage}
	\begin{minipage}{.65\textwidth}
		\tiny
		\centering
		\caption{\small State bits and their meanings.}
		\vspace{0.1cm}
		\label{my-label}
		\begin{tabular}{|l|l|l|}
			\hline
			\multicolumn{1}{|c|}{\textbf{State bits}} & \multicolumn{1}{c|}{\textbf{Name}}                                              & \multicolumn{1}{c|}{\textbf{Description}}                                                                                                               \\ \hline
			QERS                     & \begin{tabular}[c]{@{}l@{}}Query entities and \\ relation searched\end{tabular} & \begin{tabular}[c]{@{}l@{}}Whether the query source ($s$) and target ($t$) entities \\ and query relation ($r$) have been searched in KB or not.\end{tabular} \\ \hline
			SEF                        & Source Entity Found              & Whether the source entity ($s$) has been found in KB or not.   \\ \hline
			TEF                        & Target Entity Found               & Whether the target entity ($t$) has been found in KB or not        \\ \hline
			QRF                        & Query Relation Found            & Whether the query relation ($r$) has been found in KB or not     \\ \hline
			CLUE                      & Clue bit set                           & Whether the query is a clue or not.              \\ \hline
			ILO                         & Interaction Limit Over              & Whether the interaction limit is over for the query or not.        \\ \hline
			PFE                        & Path Feature extracted           & Whether path feature extraction has been done or not.          \\ \hline
			NEFS                     & Non-empty Feature set         & Whether the extracted feature set is non-empty or empty.      \\ \hline
			CPF                        & Complete path Found             & Whether the extracted path features are complete or not.      \\ \hline
			INFI                      & Inference Invoked      & Whether Inference instruction has been invoked or not.        \\ \hline
		\end{tabular}
		\normalsize
		\vspace{-0.1cm}
	\end{minipage}
\end{table*}

\begin{table*}[t]
	\vspace{-0.2cm}
	\tiny
	\centering
	\caption{\small Actions and their descriptions.}
	\vspace{0.1cm}
	\label{my-label}
	\begin{tabular}{|c|l|l|}
		\hline
		\textbf{Id} & \multicolumn{1}{c|}{\textbf{Description}}                                                                              & \multicolumn{1}{c|}{\textbf{Reward Structure~~~[condition type]}} \\ \hline
		$a_0$       & \begin{tabular}[c]{@{}l@{}}Search source ($h$), target ($t$) entities\\ and query relation ($r$) in KB.\end{tabular}      &   \(\displaystyle  r = \begin{cases} 0 &\text{if $S_t[QERS]=0$~~~[\textbf{\textit{valid}}]}\\ -1 &\text{otherwise~~~[\textbf{\textit{invalid}}]} \end{cases} \)                                      \\ \hline
		$a_1$        & \begin{tabular}[c]{@{}l@{}}Ask user to provide an example/clue for\\ query relation $r$\end{tabular}                       &       \(\displaystyle  r = \begin{cases} 0 &\text{if $S_t[ILO]=0~and~S_t[CLUE]=0~and~S_t[QERS]=1~and~S_t[QRF]=0$~~~[\textbf{\textit{valid}}]}\\ -1 &\text{otherwise~~~[\textbf{\textit{invalid}}]} \end{cases} \)      \\ \hline
		$a_2$       & \begin{tabular}[c]{@{}l@{}}Ask user to provide missing link \\ for path feature completion.\end{tabular}   &      \(\displaystyle  r = \begin{cases} 0 &\text{if $S_t[PFE]=1~and~S_t[NEFS]=1~and~S_t[ILO]=0~and~S_t[CPF]=0$~~~[\textbf{\textit{valid}}]}\\ -1 &\text{otherwise~~~[\textbf{\textit{invalid}}]} \end{cases} \)      \\ \hline
		$a_3$       & \begin{tabular}[c]{@{}l@{}}Ask user to provide the connecting link\\ for augmenting a new entity with KB.\end{tabular} &     \(\displaystyle  r = \begin{cases} -1 &\text{if $S_t[QERS]=0~or~(S_t[SEF]=1~and~S_t[TEF]=1)~or~S_t[ILO]=1$~~~[\textbf{\textit{invalid}}]}\\ 0 &\text{otherwise~~~[\textbf{\textit{valid}}]} \end{cases} \)       \\ \hline
		$a_4$       & \begin{tabular}[c]{@{}l@{}}Extract path features between source \\ ($s$) and target ($t$) entities using C-PR\end{tabular} &    \(\displaystyle  r = \begin{cases} -1 &\text{if $S_t[QERS]=0~or~S_t[PFE]=1~~~[\textbf{\textit{invalid}}]~ or~(a_4~ $executed$~, but~|\Pi_d|=0\textbf{*})$}\\ 0 &\text{otherwise~~~[\textbf{\textit{valid}}]} \end{cases} \)        \\ \hline
		$a_5$       & \begin{tabular}[c]{@{}l@{}}Store query data instance in data buffer \\and invoke prediction model for inference.\end{tabular}       &          \(\displaystyle  r = \begin{cases} 1 &\text{if $S_t[QRF]=1~and~S_t[CPF]=1$~~~[\textbf{\textit{win}}]}\\ -1 &\text{otherwise~~~[\textbf{\textit{loss}}]} \end{cases} \)   \\ \hline
	\end{tabular}
	\normalsize
	\vspace{-0.1cm}
\end{table*}

The RL model learns even after training whenever it encounters an unseen state (in testing) and thus, gets updated over time. KS is updated continuously over time as a result of the execution of LiLi and takes part in future learning. The prediction model uses lifelong learning (LL), where we transfer knowledge (parameter values) from the model for a past most similar task to help learn for the current task. Similar tasks are identified by factorizing $\mathcal{M}$ and computing a task similarity matrix $\mathcal{M}_{sim}$. Besides LL, LiLi uses $\mathcal{T}$ to identify poorly learned past tasks and acquire more clues for them to improve its skillset over time.

LiLi also uses a stack, called \textbf{\textit{Inference Stack}} ($\mathcal{IS}$) to hold query and its state information for RL. LiLi always processes stack top ($\mathcal{IS}$[top]). The clues from the user get stored in $\mathcal{IS}$ on top of the query during strategy execution and processed first. Thus, the prediction model for $r$ is learned before performing inference on query, transforming OKBC to a KBC problem. Table 1 shows the parameters of LiLi used in the following sections.

\subsection{Working of LiLi}
Given an OKBC query ($s$, $r$, $t$), we represent it as a data instance $d$. $d$ consists of $t_d$ (the query triple), $\delta_{IL}(d)$ (interaction limit set for $d$), $exp_d$ (experience list storing the transition history of MDP for $d$ in RL) and $m_d$ (mode of $d$) denoting if $d$ is `$T$' (training), `$V$' (validation), `$E$' (evaluation) or `$C$' (clue) instance and $\Pi_d$ (feature set). We denote $\Pi_d^{cp}$ ($\overline{\Pi_d^{cp}}$ ) as the set of all complete (incomplete) path features in $\Pi_d$. Given a data instance $d$, LiLi starts its initialization as follows: it sets the state as $S_0$ (based on $m_d$, explained later), pushes the query tuple ($d$, $S_0$) into $\mathcal{IS}$ and feeds $\mathcal{IS}$[top] to the RL-model for strategy formulation from $S_0$.

\textbf{Inference Strategy Formulation.}  We view solving the strategy formulation problem as learning to play an inference game, where the goal is to formulate a strategy that "\textit{makes the inference task possible}". Considering PR methods, inference is possible, \textit{iff} (1) $r$ becomes known to its KB (by acquiring clues when $r$ is unknown) and (2) path features are extracted between $s$ and $t$ (which inturn requires $s$ and $t$ to be known to KB). If these \textit{conditions are met} at the end of an episode (when strategy formulation finishes for a given query) of the game, LiLi wins and thus, it trains the prediction model for $r$ and uses it for inference.

LiLi's strategy formulation is modeled as a Markov Decision Process (MDP) with finite state ($\mathcal{S}$) and action ($\mathcal{A}$) spaces. A state $S \in \mathcal{S}$ consists of 10 binary state variables (Table 2), each of which keeps track of results of an action $a \in \mathcal{A}$ taken by LiLi and thus, records the progress in inference process made so far. $S_0$ is the initial state with all state bits set as 0. If the data instance (query) is a clue [$m_d=C$], $S_0$[CLUE] is set as 1.  $\mathcal{A}$ consists of 6 actions (Table 3). $a_0$, $a_4$, $a_5$ are \textit{processing actions} and $a_1$, $a_2$, $a_3$ are \textit{interactive actions}. Whenever $a_5$ is executed, the MDP reaches the terminal state. Given an action $a$ in state $S_t$,  if $a$ is \textit{invalid}\footnote{ ``invalid" means performing $a$ in $S_t$ is meaningless (doesn't advance reasoning) like choosing repetitive processing actions during training (by random exploration of $\mathcal{A}$).} in $S_t$ or the objective of $a$ is \textit{unsatisfied} (* marked the condition in $a_4$), RL receives a negative reward (empirically set); else receives a positive reward.\footnote{Unlike existing RL-based interactive or active learning \cite{li2016learning,woodward2017active}, the user doesn\mbox{'}t provide feedback to guide the learning process of LiLi. Rather, the RL-model uses an internal feedback mechanism to self-learn its optimal policy. This is analogous to the idea of \textit{learning by self-realization} observed in humans: whenever we try to solve a problem, we often try to formulate strategy and refine it ourselves based on whether we can derive the answer of the problem without external guidance. Likewise, here the RL-model gets feedback based on whether it is able to advance the reasoning process or not.}. We use Q-learning \cite{watkins1992q} with $\epsilon$-greedy strategy to learn the optimal policy for training the RL model. Note that, the inference strategy is independent of KB type and correctness of prediction. Thus, the RL-model is trained only once from scratch (reused thereafter for other KBs) and also, independently of the prediction model.

Sometimes the training dataset may not be enough to learn optimal policy for all $S \in \mathcal{S}$. Thus, encountering an \textit{unseen} state during test can make RL-model clueless about the action. Given a state $S$, whenever an invalid $a \in \mathcal{A} - \{a_2\}$ is chosen, LiLi remains in $S$. For $a_2$, LiLi remains in $S$ untill $\delta_{IL}>0$ (see Table 1 for $\delta_{IL}$). So, if the state remains the same for ($\delta_{IL}$+1) times, it implies LiLi has encountered a fault (\textit{an unseen state}). RL-model instantly switches to the training mode and randomly explores $\mathcal{A}$ to learn the optimal action (\textit{fault-tolerant learning}). While exploring $\mathcal{A}$, the model chooses $a_5$ only when it has tried all other $a \in \mathcal{A}$ to avoid abrupt end of  episode.

\textbf{Execution of Actions.} At any given point in time, let ($d$, $S_t$) be the current $\mathcal{IS}$[top], $\hat{a}$ is the chosen action and the current version of KS components are $G$, $\mathcal{M}$, $\mathcal{T}$ and $\Pi_{DB}$. Then, if $\hat{a}$ is \textbf{invalid} in $S_t$, LiLi \textit{only} updates $\mathcal{IS}$[top] with ($d$, $S_t$) and returns $\mathcal{IS}$[top] to RL-model. In this process, LiLi adds experience ($S_t$, $a_0$, $r$, $S_t$) in $exp_d$ and then, replaces $\mathcal{IS}$[top] with ($d$, $S_t$). If $\hat{a}$ is \textbf{valid} in $S_t$, LiLi first sets the next state $S_{t+1} = S_t$ and performs a sequence of operations $\mathcal{O}(\hat{a})$ based on $\hat{a}$ (discussed below). Unless specified, in $\mathcal{O}(\hat{a})$, LiLi always monitors $\delta_{IL}(d)$ and if $\delta_{IL}(d)$ becomes 0, LiLi sets $S_{t+1}[ILO] = 1$. Also,  whenever LiLi asks the user a query, $\delta_{IL}(d)$ is decremented by 1. Once  $\mathcal{O}(\hat{a})$ ends, LiLi updates $\mathcal{IS}$[top] with ($d$, $S_{t+1}$) and returns $\mathcal{IS}$[top] to RL-model for choosing the next action.

In $\mathcal{O}(a_0)$, LiLi searches $s$, $r$, $t$ in $G$ and sets appropriate bits in $S_{t+1}$ (see Table 2). If $r$ was unknown before and is just added to $G$ or is in the bottom $\rho$\% (see Table 1 for $\rho$) of $\mathcal{T}$\footnote{LiLi selects $\rho$\% tasks from $\mathcal{T}$ for which it has performed poorly (evaluated on validation data in our case) and acquires clue with $\beta$ probability (while processing test data). This helps in improving skills of LiLi continuously on past poorly learned tasks.}, LiLi randomly sets $QRF=0$ with probability $\beta$. If $d$ is a clue and $s, t \in G$, LiLi updates KS with triple $t_d$, where ($s$, $r$, $t$) and ($t$, $r^{-1}$, $s$) gets added to $G$ and $\mathcal{M}[r, (s,t)]$, $\mathcal{M}[r^{-1}, (s,t)]$ are set as 1.

In $\mathcal{O}(a_1)$, LiLi asks the user to provide a clue (+ve instance) for $r$ and corrupts $s$ and $t$ of the clue once at a time, to generate -ve instances by sampling nodes from $G$. These instances help in training prediction model for $r$ while executing $a_5$.

In $\mathcal{O}(a_2)$, LiLi selects an incomplete path $\pi^*$ from $\Pi_{DB}$ to formulate MLQ, such that $\pi^* \in \Pi_d$ is most frequently observed for $r$ and $sim(e_{ij}^{\pi^{*}})$ is high, given by $\pi^{*}=\underset{\pi \in \Pi_d}{\arg\max}~[log_e \Pi_{DB}[(r, \pi, e_{ij}^\pi)] * sim(e_{ij}^\pi)]$. 
Here, $sim(e_{ij}^\pi)$ denotes the contextual similarity \cite{mazumder2017context} of entity-pair $e_{ij}^\pi$. If $sim(e_{ij}^\pi)$ is high, $e_{ij}^\pi$ is more likely to possess a relation between them and so, is a good candidate for formulating MLQ. When the user does not respond to MLQ (or CLQ in $\mathcal{O}(a_3)$), the \textit{guessing mechanism} is used, which works as follows: Since contextual similarity of entity-pairs is highly correlated with their class labels \cite{mazumder2017context}, LiLi divides the similarity range [-1, 1] into three segments, using a low ($l$) and high ($h$) similarity threshold and replaces the missing link with $l$ in $\pi^{*}$ to make it complete as follows: If $h \geq sim(e_{ij}^{\pi^*}) \geq l$, $l$= ``@-\textit{LooselyRelatedTo}-@"; else if $sim(e_{ij}^{\pi^*}) \leq l$, $l$=``@-\textit{NotRelatedTo}-@"; Otherwise, $l$=``@-\textit{RelatedTo}-@". 

In $\mathcal{O}(a_3)$, LiLi asks CLQs for connecting unknown entities $s$ and/or $t$ with $G$ by selecting the most contextually relevant node (wrt $s$, $t$) from $G$, given by link
$e^*=\underset{e^\prime \in G}{\arg\max}~\mathcal{R}elv(e^\prime, s, t)$\footnote{If \# nodes in $G$ is very large, a candidate set for $e^\prime$ is sampled for computing $e^*$.}. We adopt the contextual relevance idea in \cite{mazumder2017context} which is computed using word embedding \cite{mikolov2013distributed}\footnote{Although $s$ and $t$ may be unknown to $G$, to avoid unnecessary complexity, we assume that LiLi has access to embedding vectors for all entities (known and unknown) in our datasets. In practice, we can update the embedding model continuously by fetching documents from the Web for unknown entities.} 

In $\mathcal{O}(a_4)$, LiLi extracts path features $\Pi_d$ between ($s$, $t$) and updates $\Pi_{DB}$ with incomplete features from $\Pi_d$. LiLi always trains the prediction model with complete features $\Pi_d^{cp}$ and once $|\Pi_d^{cp}|=\delta_\pi$ or $\delta_{IL}(d)=0$, LiLi stops asking MLQs. Thus, in both $\mathcal{O}(a_4)$ and $\mathcal{O}(a_2)$, LiLi always monitors $\Pi_d$ to check for the said requirements and sets $CPF$ to control interactions.

In $\mathcal{O}(a_5)$, if LiLi wins the episode, it adds $d$ in one of data buffers $\mathcal{D}_{m_d}$ based on its mode $m_d$. E.g., if $m_d=T$ or $C$, $d$ is used for training and added to $\mathcal{D}_{tr}$. Similarly validation buffer $\mathcal{D}_{val}$ and evaluation buffer $\mathcal{D}_{eval}$ are populated. If $|\mathcal{D}_{eval}| > 0$, LiLi invokes the prediction model for $r$\footnote{We invoke the prediction model only when all instances for $r$ are populated in the data buffers to enable batch processing.}.

\textbf{Lifelong Relation Prediction.}  Given a relation $r$, LiLi uses $\mathcal{D}_{tr}$ and $\mathcal{D}_{val}$ (see $\mathcal{O}(a_5)$) to train a prediction model (say, $\mathcal{F}_r$) with parameters $\Theta_r$. For a unknown $r$, the clue instances get stored in $\mathcal{D}_{tr}$ and $|\mathcal{D}_{val}| = 0$. Thus, LiLi populates $\mathcal{D}_{val}$ by taking 10\% (see \S4) of the instances from $\mathcal{D}_{tr}$ and starts the training. For $d \in \mathcal{D}_{tr}$, LiLi uses a LSTM \cite{hochreiter1997long} to compose the vector representation of each feature $\pi \in \Pi_d$ as $v_\pi$ and vector representation of $r$ as $v_r$. Next, LiLi computes the prediction value, $\mathbb{P}(r|s,t)$ as sigmoid of the mean cosine similarity of all features and $r$, given by $\mathbb{P}(r|s,t)=sigmoid(\frac{1}{|\Pi_d|} \underset{\pi \in \Pi_d}{\sum}~cos(v_r, v_\pi)$) and maximize the log-likelihood of $\mathcal{D}_{tr}$ for training. Once $\mathcal{F}_r$ is trained, LiLi updates $\mathcal{T}$[$r$] using $\mathcal{D}_{val}$. We also train an inverse model for $r$, $\mathcal{F}_{r^{-1}}$ by \textit{reversing} the path features in $\mathcal{D}_{tr}$ and $\mathcal{D}_{val}$ which help in lifelong learning (discussed below). Unlike \cite{neelakantan2015compositional,das2016chains}, while predicting the label for $d \in \mathcal{D}_{eval}$, we compute a \textit{relation-specific} prediction threshold $\mu_r$ corresponding to $r$ using $\mathcal{D}_{val}$ as: $\mu_r=\frac{1}{2}[\mu_r^{+}+\mu_r^{-}]$  and infer $d$ as +ve if $\mathbb{P}(r|h,t) \geq \mu_r$ and -ve otherwise. Here, $\mu_r^{+}$ ($\mu_r^{-}$) is the mean prediction value for all +ve (-ve) examples in $\mathcal{D}_{val}$.

Models trained on a few examples (e.g., clues acquired for unknown $r$) with randomly initialized weights often perform poorly due to underfitting. Thus, we transfer knowledge (weights) from the past most similar (wrt $r$) task in a lifelong learning manner~\cite{ChenAndLiubook2016}.
LiLi uses $\mathcal{M}$ to find the past most similar task for $r$ as follows: LiLi computes trancated SVD of $\mathcal{M}$ as $\mathcal{M}_k=U_k \Sigma_k V^{T}_k$ and then, the similarity matrix $\mathcal{M}_{sim}= U_k \Sigma_k \Sigma_k^T U_k^{T}$.  $\mathcal{M}_{sim}(r,r^\prime)$ provides the similarity between relations $r$ and $r^\prime$ in $G$. Thus, LiLi chooses a source relation $r_s=\underset{r^\prime \in \mathcal{R}_{tr}}{\arg\max}~\mathcal{M}_{sim}(r, r^\prime)$ to transfer weights. Here, $\mathcal{R}_{tr}$ is the set of all $r$ and $r^{-1}$ for which LiLi has already learned a prediction model. Now, if $\mathcal{M}_{sim}(r, r_s) \leq 0$ or $\mathcal{R}_{tr}=\phi$, LiLi randomly initializes the weights $\Theta_r$ for $\mathcal{F}_r$ and proceeds with the training. Otherwise, LiLi uses $\Theta_{r_s}$ as initial weights and fine-tunes  $\Theta_r$ with a low learning rate.

\textbf{A Running Example.} Considering the example shown in Figure 1, LiLi works as follows: first, LiLi executes $a_0$ and detects that the source entity ``\textit{Obama}" and query relation ``\textit{CitizenOf}" are \textit{unknown}. Thus, LiLi executes $a_1$ to acquire clue (SF1) for ``\textit{CitizenOf}" and pushes the clue (+ve example) and two generated -ve examples into $\mathcal{IS}$. Once the clues are processed and a prediction model is trained for ``\textit{CitizenOf}" by formulating separate strategies for them, LiLi becomes aware of ``\textit{CitizenOf}". Now, as the clues have already been popped from $\mathcal{IS}$, the query becomes $\mathcal{IS}[top]$ and the strategy formulation process for the query resumes. Next, LiLi asks user to provide a connecting link for ``\textit{Obama}" by performing $a_3$. Now, the query entities and relation being known, LiLi enumerates paths between ``\textit{Obama}" and ``\textit{USA}" by performing $a_4$. Let an extracted path be ``$Obama-BornIn \rightarrow Honolulu-@-?-@\rightarrow Hawaii-StateOf\rightarrow USA$" with missing link between ($Honolulu$, $Hawaii$). LiLi asks the user to fill the link by performing $a_2$ and then, extracts the complete feature ``$BornIn \rightarrow CapitalOfState \rightarrow StateOf$". The feature set is then fed to the prediction model and inference is made as a result of $a_5$. Thus, the formulated inference strategy is: ``$\langle a_0, a_1, a_3, a_4, a_2, a_5\rangle$".

\section{Experiments}
We now evaluate LiLi in terms of its predictive performance and strategy formulation abilities.

\begin{table}[t]
		\tiny
		\vspace{-0.2cm}
		\caption{\small Dataset statistics [\textit{kwn} = known, \textit{unk} = unknown]}
		\vspace{0.15cm}
		\label{my-label}
		\begin{tabular}{|l|c|c|c|c|c|c|}
			\hline
			& \multicolumn{3}{c|}{\textbf{Freebase (FB)}} & \multicolumn{3}{c|}{\textbf{WordNet (WN)}} \\ \hline
			\# Rels ($\mathcal{K}_{org}$ /$\mathcal{K}_{B}$ )           & \multicolumn{3}{c|}{1,345 / 1,248}          & \multicolumn{3}{c|}{18 / 14}               \\ \hline
			\# Entities ($\mathcal{K}_{org}$ /$\mathcal{K}_{B}$)            & \multicolumn{3}{c|}{13, 871 / 12, 306}      & \multicolumn{3}{c|}{13, 595 / 12, 363}     \\ \hline
			\# Triples ($\mathcal{K}_{org}$ /$\mathcal{K}_{B}$)             & \multicolumn{3}{c|}{854, 362 / 529,622}     & \multicolumn{3}{c|}{107, 146 / 58, 946}    \\ \hline
			\# Test Rels (\textit{kwn} / \textit{unk})           & \multicolumn{3}{c|}{50 (38 / 12)}           & \multicolumn{3}{c|}{18 (14 / 4)}           \\ \hline
			Avg. \# train / valid & \multicolumn{3}{c|}{}   & \multicolumn{3}{c|}{}   \\ hline 
			/ test instances/rel. & \multicolumn{3}{c|}{1715 / 193 / 557}       & \multicolumn{3}{c|}{994 / 109 / 326}       \\ \hline
			Entity statistics (Avg. \%)                 & train         & valid        & test         & train        & valid        & test         \\ \hline
			only source ($s$) \textit{unk}                        & 15.5         & 15.8        & 15.6        & 12.4        & 10.4        & 19.0        \\ \hline
			only target ($t$) \textit{unk}                         & 13.0         & 12.7        & 13.4        & 14.2        & 15.6        & 13.8        \\ \hline
			both $s$  and $t$ \textit{unk}              & 2.9         & 3.3         & 2.8         & 3.6         & 3.6         & 6.2         \\ \hline
		\end{tabular}
		\normalsize
		\vspace{-0.3cm}
\end{table}

\noindent \textbf{Data:}
We use two standard datasets (see Table 4): (1) Freebase FB15k\footnote{ https://everest.hds.utc.fr/doku.php?id=en:smemlj12}, and (2) WordNet$^{12}$.  Using each dataset, we build a fairly large graph and use it as the original KB ($\mathcal{K}_{org}$) for evaluation. We also augment $\mathcal{K}_{org}$ with inverse triples ($t$, $r^{-1}$, $s$) for each ($s$, $r$, $t$) following existing KBC methods. 

\begin{table*}[t]
	\tiny
	\centering
	\caption{\small Inference strategies formulated by LiLi (ordered by frequency).}
	\vspace{0.2cm}
	\label{my-label}
	\begin{tabular}{|l|l|l|l|l|}
		\hline
		\multicolumn{1}{|c|}{\textbf{$\mathbf{\delta_{IL}=0, \delta_\pi=3}~$ {[}C: 0.47{]}}} & $\langle a_0, a_1, a_4, a_2, a_2, a_5 \rangle$ & \multicolumn{1}{c|}{\textbf{$\mathbf{\delta_{IL}=5, \delta_\pi=1}~$ {[} C: 1.0{]}}} & \multicolumn{2}{c|}{\textbf{$\mathbf{\delta_{IL}=5, \delta_\pi=3}~$~ {[}C: 1.0{]}}}  \\ \cline{1-1} \cline{3-5}
		$\langle a_0, a_4, a_5\rangle$                                     & $\langle a_0, a_3, a_4, a_2, a_2, a_5 \rangle$ & $\langle a_0, a_4, a_2, a_5 \rangle $                            & $\langle a_0, a_4, a_2, a_2, a_2, a_5\rangle $      & $\langle a_0, a_3, a_4, a_5\rangle $                     \\ \cline{1-1}
		\multicolumn{1}{|c|}{\textbf{$\mathbf{\delta_{IL}=1, \delta_\pi=3}~$ {[} C: 0.97{]}}} & $\langle a_0, a_4, a_5 \rangle $               & $\langle a_0, a_1, a_4, a_2, a_5 \rangle $                       & $\langle a_0, a_4, a_2, a_5\rangle $                & $\langle a_0, a_1, a_4, a_5\rangle $                     \\ \cline{1-1}
		$\langle a_0, a_4, a_2, a_5\rangle$                                & $\langle a_0, a_3, a_4, a_2, a_5 \rangle $     & $\langle a_0, a_4, a_5 \rangle $                                 & $\langle a_0, a_4, a_2, a_2, a_5\rangle $           & $\langle a_0, a_1, a_3, a_4, a_2, a_2, a_2, a_5\rangle $ \\ 
		$\langle a_0, a_1, a_4, a_5\rangle$                                & $\langle a_0, a_1, a_4, a_2, a_5 \rangle $     & $\langle a_0, a_3, a_4, a_2, a_5 \rangle $                       & $\langle a_0, a_1, a_4, a_2, a_2, a_2, a_5\rangle $ & $\langle a_0, a_1, a_3, a_4, a_2, a_2, a_5\rangle $      \\ 
		$\langle a_0, a_3, a_4, a_5\rangle$                                & $\langle a_0, a_3, a_4, a_5 \rangle $          & $\langle a_0, a_1, a_4, a_5 \rangle $                            & $\langle a_0, a_4, a_5\rangle $                     & $\langle a_0, a_1, a_3, a_4, a_2, a_5\rangle $           \\ 
		$\langle a_0, a_4, a_5\rangle$                                     & $\langle a_0, a_1, a_3, a_4, a_2, a_5 \rangle$ & $\langle a_0, a_3, a_4, a_5 \rangle $                            & $\langle a_0, a_3, a_4, a_2, a_2, a_2, a_5\rangle $ & $\langle a_0, a_3, a_1, a_4, a_2, a_2, a_5\rangle $      \\ \cline{1-1}
		\multicolumn{1}{|c|}{\textbf{$\mathbf{\delta_{IL}=3, \delta_\pi=3}~$~{[}C: 1.0{]}}}  & $\langle a_0, a_3, a_1, a_4, a_2, a_5 \rangle$ & $\langle a_0, a_3, a_1, a_4, a_2, a_5 \rangle $                  & $\langle a_0, a_3, a_4, a_2, a_5\rangle $           & $\langle a_0, a_3, a_1, a_4, a_2, a_2, a_2, a_5\rangle $ \\ \cline{1-1}
		$\langle a_0, a_4, a_2, a_2, a_2, a_5 \rangle$                     & $\langle a_0, a_1, a_4, a_5 \rangle$           & $\langle a_0, a_1, a_3, a_4, a_2, a_5 \rangle $                  & $\langle a_0, a_3, a_4, a_2, a_2, a_5\rangle $      & $\langle a_0, a_3, a_1, a_4, a_5\rangle $                \\ 
		$\langle a_0, a_4, a_2, a_5 \rangle $                              & $\langle a_0, a_3, a_1, a_4, a_5 \rangle$      & $\langle a_0, a_3, a_1, a_4, a_5 \rangle $                       & $\langle a_0, a_1, a_4, a_2, a_2, a_5\rangle $      & $\langle a_0, a_3, a_1, a_4, a_2, a_5\rangle $           \\ 
		$\langle a_0, a_4, a_2, a_2, a_5 \rangle $                         & $\langle a_0, a_1, a_3, a_4, a_5 \rangle$      & $\langle a_0, a_1, a_3, a_4, a_5 \rangle $                       & $\langle a_0, a_1, a_4, a_2, a_5\rangle $           & $\langle a_0, a_1, a_3, a_4, a_5\rangle $                \\ \hline
	\end{tabular}
	\normalsize
	\vspace{-0.25cm}
\end{table*}

\begin{table*}[t]
	\scriptsize
	\centering
	\caption{\small Comparison of predictive performance of \textit{various versions} of LiLi [\textit{kwn} = known, \textit{unk} = unknown, \textit{all} = overall].}
	\vspace{0.2cm}
	\label{my-label}
	\begin{tabular}{|c|l|c|c|c|c|c|c|c|c|c|c|c|c|}
		\hline
		\multirow{2}{*}{KB} & \multicolumn{1}{c|}{\multirow{2}{*}{\begin{tabular}[c]{@{}c@{}}Test \\ Rel\\ type\end{tabular}}} & \multicolumn{6}{c|}{Avg.  +ve F1 Score}                                                                                                                                                                                                                                                                                                  & \multicolumn{6}{c|}{Avg. MCC}                                                                                                                                                                                                                                                                                                            \\ \cline{3-14} 
		& \multicolumn{1}{c|}{}                                                                                     & \begin{tabular}[c]{@{}c@{}} Single\end{tabular} & \begin{tabular}[c]{@{}c@{}}Sep\end{tabular} & \begin{tabular}[c]{@{}c@{}}F-th\end{tabular} & \begin{tabular}[c]{@{}c@{}}BG\end{tabular} & \begin{tabular}[c]{@{}c@{}}w/o \\ PTS\end{tabular} & LiLi    & \begin{tabular}[c]{@{}c@{}}Single\end{tabular} & \begin{tabular}[c]{@{}c@{}}Sep\end{tabular} & \begin{tabular}[c]{@{}c@{}} F-th\end{tabular} & \begin{tabular}[c]{@{}c@{}}BG\end{tabular} & \begin{tabular}[c]{@{}c@{}}w/o \\ PTS\end{tabular} & LiLi   \\ \hline
		\multirow{3}{*}{FB}          & \textit{kwn}                                                                                            & 0.3796                                                          & 0.5741                                                       & 0.5069                                                        & 0.5643                                                      & 0.5547                                                           & \textbf{0.5859} & 0.0937                                                          & 0.2638                                                       & 0.2382                                                        & 0.2443                                                      & 0.2573                                                           & \textbf{0.2763} \\ \cline{2-14} 
		& \textit{unk}                                                                                          & 0.5477                                                          & 0.5425                                                       & 0.4876                                                        & 0.5398                                                      & 0.5421                                                           & \textbf{0.5567} & \textbf{0.2175}                                                 & 0.1752                                                       & 0.1802                                                        & 0.1664                                                      & 0.1748                                                           & 0.2119          \\ \cline{2-14} 
		& \textit{\textbf{all}}                                                                                 & 0.4199                                                          & 0.5665                                                       & 0.5023                                                        & 0.5584                                                      & 0.5517                                                           & \textbf{0.5789} & 0.1234                                                          & 0.2425                                                       & 0.2243                                                        & 0.2256                                                      & 0.2375                                                           & \textbf{0.2609} \\ \hline
		\multirow{3}{*}{WN}          & \textit{kwn}                                                                                            & 0.3846                                                          & 0.5851                                                       & 0.5817                                                        & 0.5554                                                      & 0.6083                                                           & \textbf{0.6343} & 0.2494                                                          & 0.3838                                                       & 0.3603                                                        & 0.2980                                                      & \textbf{0.4159}                                                  & 0.4096          \\ \cline{2-14} 
		& \textit{unk}                                                                                          & 0.5732                                                          & 0.5026                                                       & 0.5861                                                        & 0.5694                                                      & 0.5539                                                           & \textbf{0.5871} & 0.3348                                                          & 0.2501                                                       & 0.3123                                                        & 0.3148                                                      & 0.2667                                                           & \textbf{0.3387} \\ \cline{2-14} 
		& \textit{\textbf{all}}                                                                                 & 0.4265                                                          & 0.5668                                                       & 0.5827                                                        & 0.5586                                                      & 0.5962                                                           & \textbf{0.6238} & 0.2684                                                          & 0.3541                                                       & 0.3496                                                        & 0.3017                                                      & 0.3828                                                           & \textbf{0.3939} \\ \hline
	\end{tabular}
	\normalsize
	\vspace{-0.3cm}
\end{table*}

\noindent 
\textbf{Parameter Settings.} Unless specified, the \textit{empirically} set parameters (see Table 1) of LiLi are: $\alpha=0.8$, $\gamma=0.9$, $\delta_{IL}=5$, $\delta_{\pi}=3$, $\eta_{\pi}=7$, $\eta_{w}=20$, $l=0.07$, $h=0.2$, $k=300$, $\beta=0.5$, $\rho=25\%$. For training RL-model with $\epsilon$-greedy strategy, we use $\epsilon_{start}=1.0$, $\epsilon_{end}=0.1$, pre-training steps=50000. We used Keras deep learning library to implement and train the prediction model. We set batch-size as 128, max. training epoch as 150, dropout as 0.2, hidden units and embedding size as 300 and learning rate as 5e-3 which is reduced gradually on plateau with factor 0.5 and patience 5. Adam optimizer and early stopping were used in training. We also shuffle $\mathcal{D}_{tr}$ in each epoch and adjust class weights inversely proportional to class frequencies in $\mathcal{D}_{tr}$.

\noindent 
\textbf{Labeled Dataset Generation and Simulated User Creation.} We create a simulated user for each KB to evaluate LiLi\footnote{Crowdsourced-based training and evaluation is expensive and time consuming as user-interaction is needed in training.}. We create the labeled datasets, the simulated user’s knowledge base ($\mathcal{K}_{u}$), and the base KB ($\mathcal{K}_{B}$) from $\mathcal{K}_{org}$. $\mathcal{K}_{B}$ used as the initial KB graph ($G_0$) of LiLi.

We followed \cite{mazumder2017context} for labeled dataset generation.  For Freebase, we found 86 relations with $\geq 1000$ triples and randomly selected 50 from various domains. We randomly shuffle the list of 50 relations, select 25\% of them as \textit{unknown} relations and consider the rest (75\%) as \textit{known} relations. For each \textit{known} relation $r$, we randomly shuffle the list of distinct triples for $r$, choose 1000 triples and split them into 60\% training, 10\% validation and 20\% test. Rest 10\% along with the leftover (not included in the list of 1000) triples are added to $\mathcal{K}_{u}$. For each \textit{unknown} relation $r$, we remove all triples of $r$ from $\mathcal{K}_{org}$ and add them to $\mathcal{K}_{u}$. In this process, we also randomly choose 20\% triples as test instances for unknown $r$ which are excluded from $\mathcal{K}_{u}$. Note that, now $\mathcal{K}_{u}$ has at least 10\% of chosen triples for each $r$ (\textit{known} and \textit{unknown}) and so, user is always able to provide clues for both cases. For each labeled dataset, we randomly choose 10\% of the entities present in dataset triples, remove triples involving those entities from $\mathcal{K}_{org}$ and add to $\mathcal{K}_{u}$. At this point, $\mathcal{K}_{org}$ gets reduced to $\mathcal{K}_{B}$ and is used as $G_0$ for LiLi. The dataset stats in Table 4 shows that the base KB (60\% triples of $\mathcal{K}_{org}$) is \textit{highly sparse} (compared to original KB) which \textit{makes the inference task much harder}. WordNet dataset being small, we select all 18 relations for evaluation and create labeled dataset, $\mathcal{K}_{u}$ and $\mathcal{K}_{B}$ following Freebase. Although the user may provide clues 100\% of the time, it often cannot respond to MLQs and CLQs (due to lack of required triples/facts). Thus, we further enrich $\mathcal{K}_{u}$ with external KB triples\footnote{Due to fair amount of entity overlapping, we choose NELL for enriching $\mathcal{K}_{u}$ in case of Freebase and ConceptNet for enriching $\mathcal{K}_{u}$ in case of WordNet.}. 

Given a relation $r$ and an observed triple ($s$, $r$, $t$) in training or testing, the pair ($s$, $t$) is regarded as a +ve instance for $r$. Following \cite{wang2016knowledge}, for each +ve instance ($s$, $t$), we generate two negative ones, one by randomly corrupting the source $s$, and the other by corrupting the target $t$. Note that, the test triples are not in $\mathcal{K}_{B}$ or $\mathcal{K}_{u}$ and none of the -ve instances overlap with the +ve ones.

\noindent \textbf{Baselines.}  As none of the existing KBC methods can solve the OKBC problem, we choose various versions of LiLi as baselines.\\
\indent \textbf{Single}: Version of LiLi where we train a single prediction model $\mathcal{F}$ for all test relations.\\
\indent \textbf{Sep}: We do not transfer (past learned) weights for initializing $\mathcal{F}_r$, i.e., we \textbf{disable} LL.\\
\indent \textbf{F-th):} Here, we use a \textit{ fixed} prediction threshold 0.5 instead of relation-specific threshold $\mu_r$.\\
\indent \textbf{BG:} The missing or connecting links (when the user does not respond) are filled with ``@-\textit{RelatedTo}-@" blindly, no guessing mechanism.\\
\indent \textbf{w/o PTS:} LiLi does not ask for additional clues via past task selection for skillset improvement.

\noindent \textbf{Evaluation Metrics.} To evaluate the strategy formulation ability, we introduce a measure called \textit{Coverage}($C$), defined as the fraction of total query data instances, for which LiLi has successfully formulated strategies that lead to winning. If LiLi wins on all episodes for a given dataset, $C$ is 1.0.  To evaluate the predictive performance, we use Avg. MCC and avg. +ve F1 score.

\subsection{Results and Analysis}
\textbf{Evaluation-I: Strategy Formulation Ability.} Table 5 shows the list of inference strategies formulated by LiLi for various $\delta_{IL}$ and $\delta_\pi$, which control the strategy formulation of LiLi. When $\delta_{IL}=0$, LiLi cannot interact with user and works like a closed-world method. Thus, $C$ drops significantly (0.47). When $\delta_{IL}=1$, i.e. with only one interaction per query, LiLi acquires knowledge well for instances where either of the entities or relation is unknown. However, as one unknown entity may appear in multiple test triples, once the entity becomes known, LiLi doesn’t need to ask for it again and can perform inference on future triples causing significant increase in $C$ (0.97). When $\delta_{IL}=3$, LiLi is able to perform inference on all instances and $C$ becomes 1. For $\delta_\pi=1$, LiLi uses $a_2$ only once (as only one MLQ satisfies $\delta_\pi$) compared to $\delta_\pi=3$. In summary, LiLi’s RL-model can effectively formulate query-specific inference strategies (based on specified parameter values). 

\textbf{Evaluation-II: Predictive Performance.} Table 6 shows the comparative performance of LiLi with baselines. To judge the overall improvements, we performed paired t-test considering +ve F1 scores on each relation as paired data. Considering both KBs and all relation types, LiLi outperforms Sep with $p < 0.1$. If we set $\beta=0.05$ (training with \textit{very few} clues), LiLi outperforms Sep with $p < 0.05$ on Freebase considering MCC. Thus, the lifelong learning mechanism is effective in transferring helpful knowledge. Single model performs better than Sep for unknown relations due to the sharing of knowledge (weights) across tasks. However, for known relations, performance drops because, as a new relation arrives to the system, old weights get corrupted and catastrophic forgetting occurs. For unknown relations, as the relations are evaluated just after training, there is no chance for catastrophic forgetting. The performance improvement ($p < 0.05$) of LiLi over F-th on Freebase signifies that the relation-specific threshold $\mu_r$ works better than fixed threshold 0.5 because, if all prediction values for test instances lie above (or below) 0.5, F-th predicts all instances as +ve (-ve) which degrades its performance. Due to the utilization of contextual similarity (highly correlated with class labels) of entity-pairs, LiLi’s guessing mechanism works better ($p < 0.05$) than blind guessing (BG). The past task selection mechanism of LiLi also improves its performance over w/o PTS, as it acquires more clues during testing for poorly performed tasks (evaluated on validation set). For Freebase, due to a large number of past tasks [9 (25\% of 38)], the performance difference is more significant ($p < 0.01$). For WordNet, the number is relatively small [3 (25\% of 14)] and hence, the difference is not significant.

\textbf{Evaluation-III: User Interaction vs. Performance.} Table 7 shows the results of LiLi by varying clue acquisition rate ($\beta$). We use Freebase for tuning $\beta$ due to its higher number of unknown test relations compared to WordNet. LiLi’s performance improves significantly as it acquires more clues from the user. The results on $\beta=0.5$ outperforms ($p < 0.05$) that on $\beta=0.05$. Table 8 shows the results of LiLi on user responses to MLQ’s and CLQ’s. Answering MLQ’s and CLQ’s is very hard for simulated users (unlike crowd-sourcing) as often $\mathcal{K}_u$ lacks the required triple. Thus, we attempt to analyze how the performance is effected if the user does not respond at all. The results show a clear trend in overall performance improvement when the user responds. However, the improvement is not significant as the simulated user’s query satisfaction rate (1\% MLQs and 10\% CLQs) is very small. But, the analysis shows the effectiveness of LiLi’s \textit{guessing mechanism} and continual learning ability that help in achieving avg. +ve F1 of 0.57 and 0.62 on FB and WN respectively with minimal participation of the user. 

\begin{table}[t]
	\small
	\centering
	\caption{\small LiLi's performance on FB by varying $\beta$.}
	\vspace{0.1cm}
	\label{my-label}
	\begin{tabular}{|l|c|c|c|}
		\hline
		\multicolumn{1}{|c|}{\multirow{2}{*}{\textbf{\begin{tabular}[c]{@{}c@{}}Rel\\ Type\end{tabular}}}} & \textbf{$\mathbf{\beta=0.05}$} & \textbf{$\mathbf{\beta=0.25}$} & \textbf{$\mathbf{\beta=0.5}$} \\ \cline{2-4} 
		\multicolumn{1}{|c|}{}                                                                             & \textbf{F(+)}                  & \textbf{F(+)}                  & \textbf{F(+)}                 \\ \hline
		\textit{known}                                                                                     & 0.5796                         & 0.5820                         & \textbf{0.5859}               \\ \hline
		\textit{unknown}                                                                                   & 0.5231                         & 0.5414                         & \textbf{0.5567}               \\ \hline
		\textit{\textbf{overall}}                                                                          & 0.5660                         & 0.5722                         & \textbf{0.5789}               \\ \hline
	\end{tabular}
	\normalsize
	\vspace{-0.3cm}
\end{table}  

\begin{table}[t]
	\small
	\centering
	\caption{\small Performance of LiLi on user's responses.}
	\vspace{0.1cm}
	\label{my-label}
	\begin{tabular}{|l|l|c|c|c|c|}
		\hline
		\multicolumn{1}{|c|}{\multirow{2}{*}{\textbf{KB}}} & \multicolumn{1}{c|}{\multirow{2}{*}{\textbf{\begin{tabular}[c]{@{}c@{}}Rel\\ Type\end{tabular}}}} & \multicolumn{2}{c|}{\textbf{\begin{tabular}[c]{@{}c@{}}No Response to \\ CLQs and MLQs\end{tabular}}} & \multicolumn{2}{c|}{\textbf{\begin{tabular}[c]{@{}c@{}}Response to \\ CLQs and MLQs\end{tabular}}} \\ \cline{3-6} 
		\multicolumn{1}{|c|}{}                             & \multicolumn{1}{c|}{}                                                                             & \textbf{F(+)}                                     & \textbf{MCC}                                      & \textbf{F(+)}                                    & \textbf{MCC}                                    \\ \hline
		\multirow{3}{*}{FB}                                & \textit{known}                                                                                    & 0.5823                                            & \textbf{0.2775}                                   & \textbf{0.5859}                                  & 0.2763                                          \\ \cline{2-6} 
		& \textit{unknown}                                                                                  & 0.5529                                            & 0.2049                                            & \textbf{0.5567}                                  & \textbf{0.2119}                                 \\ \cline{2-6} 
		& \textit{\textbf{overall}}                                                                         & 0.5753                                            & 0.2601                                            & \textbf{0.5789}                                  & \textbf{0.2609}                                 \\ \hline
		\multirow{3}{*}{WN}                                & \textit{known}                                                                                    & 0.5990                                            & 0.3590                                            & \textbf{0.6343}                                  & \textbf{0.4096}                                 \\ \cline{2-6} 
		& \textit{unknown}                                                                                  & \textbf{0.5952}                                   & \textbf{0.3457}                                   & 0.5871                                           & 0.3387                                          \\ \cline{2-6} 
		& \textit{\textbf{overall}}                                                                         & 0.5982                                            & 0.3561                                            & \textbf{0.6238}                                  & \textbf{0.3939}                                 \\ \hline
	\end{tabular}
	\normalsize
	\vspace{-0.3cm}
\end{table}

\section{Conclusion}
\label{sec:length}
\vspace{-0.2cm}
In this paper, we are interested in building a generic engine for continuous knowledge learning in human-machine conversations. We first showed that the problem underlying the engine can be formulated as an open-world knowledge base completion (OKBC) problem. We then proposed an lifelong interactive learning and inference (LiLi) approach to solving the OKBC problem. OKBC is a generalization of KBC. LiLi solves the OKBC problem by first formulating a query-specific inference strategy using RL and then executing it to solve the problem by interacting with the user in a lifelong learning manner. Experimental results showed the effectiveness of LiLi in terms of both predictive quality and strategy formulation ability. We believe that a system with the LiLi approach can serve as a knowledge learning engine for conversations. Our future work will improve LiLi to make more accurate. 

\section*{Acknowledgments}

This work was supported in part by National Science Foundation (NSF) under grant no.~IIS-1407927 and IIS-1650900, and a gift from Huawei Technologies Co Ltd.

\bibliography{naaclhlt2018}
\bibliographystyle{acl_natbib}
\end{document}